\documentclass{article}

\usepackage{arxiv}

\usepackage[utf8]{inputenc} 
\usepackage[T1]{fontenc}    
\usepackage{hyperref}       
\usepackage{url}            
\usepackage{booktabs}       
\usepackage{amsfonts}       
\usepackage{nicefrac}       
\usepackage{microtype}      
\usepackage{lipsum}         
\usepackage{graphicx}
\usepackage[natbibapa]{apacite}
\usepackage{doi}

\usepackage{subcaption}
\usepackage{mathtools}
\usepackage{xcolor}
\usepackage{multicol}

\newcommand{\fig}[1]{Fig.~\ref{#1}}

\usepackage{wrapfig}

\usepackage{dsfont}
\usepackage{amsmath}

\title{The Alignment Ceiling: Objective Mismatch in \\Reinforcement Learning from Human Feedback
}

\renewcommand{\shorttitle}{Objective Mismatch in RLHF}

\author{
  Nathan Lambert\\
  Allen Institute for AI \\
  Berkeley, CA, USA\\
  \texttt{nathanl@allenai.org} \\
   \And
  Roberto Calandra \\
  TU Dresden \\
  Dresden, Germany \\
  \texttt{roberto.calandra@tu-dresden.de} \\
}

\begin{document}
\maketitle

\begin{abstract}
Reinforcement learning from human feedback (RLHF) has emerged as a powerful technique to make large language models (LLMs) more capable in complex settings.
RLHF proceeds as collecting human preference data, training a reward model on said data, and optimizing a base ML model with respect to said reward for extrinsic evaluation metrics (e.g. MMLU, GSM8k).
RLHF relies on many assumptions about how the various pieces fit together, such as a reward model capturing human preferences and an RL optimizer extracting the right signal from a reward model.
As the RLHF process involves many distinct design decisions, it is easy to assume that multiple processes are correlated and therefore numerically linked.
This apparent correlation is often not true, where reward models are easily overoptimized or RL optimizers can reduce performance on tasks not modeled in the data.
Notable manifestations of models trained with imperfect RLHF systems are those that are prone to refusing basic requests for safety reasons or appearing lazy in generations.
As chat model evaluation becomes increasingly nuanced, the reliance on a perceived link between reward model training, RL scores, and downstream performance drives these issues, which we describe as an \textit{objective mismatch}.
In this paper, we illustrate the causes of this issue, reviewing relevant literature from model-based reinforcement learning, and argue for solutions.
By solving objective mismatch in RLHF, the ML models of the future will be more precisely aligned to user instructions for both safety and helpfulness.
\end{abstract}


\section{Introduction}
Reinforcement learning from human feedback (RLHF) is a powerful tool for integrating qualitative values into large machine learning models~\citep{christiano2017deep, ouyang2022training, bai2022training} that are used in popular consumer apps such as ChatGPT and Midjourney.
RLHF was popularized with its use to integrate human values into large language models (LLMs) for aligning chat tools~\citep{chatgpt,team2023gemini}.
RLHF has become an important technique in the process of making models better at responding to user requests, often referred to as instruction-tuned, steerable, aligned, or chat-tuned.

RLHF methods typically operate in a multi-step process on top of a base language model, first learning a model of human preferences that acts as a reward function, and second using this model within a reinforcement learning (RL) loop.
These two steps are often executed independently, with a reward model (RM) being trained on human preference data and then the RL optimizer is used to extract maximum information from the RM into the base model.
This multi-step process induces challenges~\citep{schulman2023proxy} -- even the most popular RLHF models include weaknesses such as \texttt{llama-2-70b-chat-hf}'s propensity to refuse vanilla requests on safety grounds~\citep{rottger2023xstest} or a version of ChatGPT documented officially as having ``cases of laziness''~\citep{openai_2024}.
Colloquially, these issues fall under the potential banner of ``too much RLHF.''
These failures are signs of the current limitations of RLHF, where even with positive signals in training of each individual module, the resulting model can have unintended behaviors.
\begin{figure}[t]
    \centering
    \includegraphics[width=0.7\textwidth]{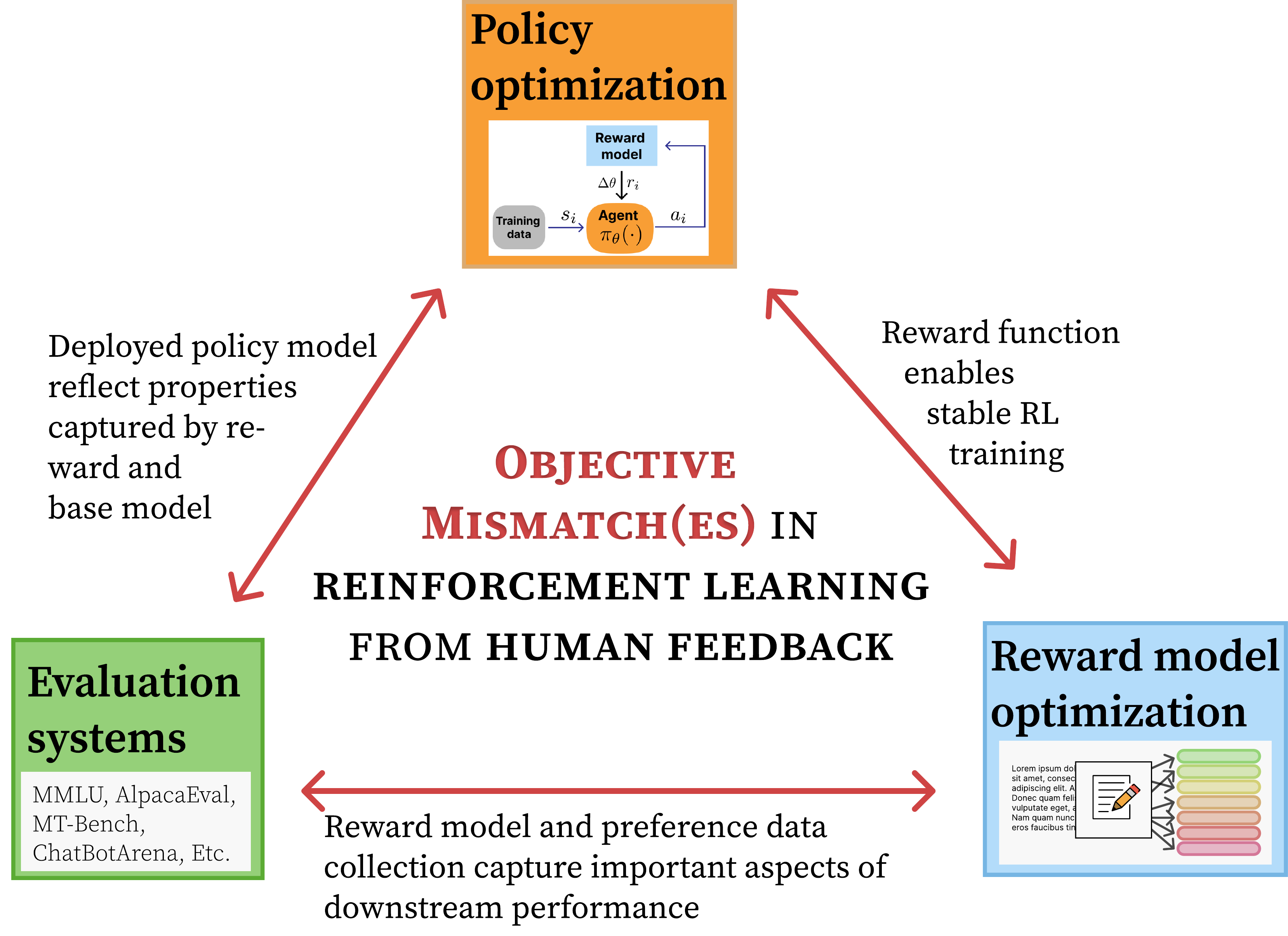}
    \caption{The three links causing objective mismatch in RLHF: Reward model training $\leftrightarrow$ policy model training, reward model training $\leftrightarrow$ evaluation tools, and policy model training $\leftrightarrow$ evaluation tools, as discussed in Sec.~\ref{sec:origins}.
    }
    \label{fig:mismatch_detailed}
\end{figure}

In this paper, we detail and argue for solving a fundamental challenge in modern RLHF learning schemes -- \textit{objective mismatch} -- in order to mitigate these issues. 
In RLHF, three important parts of training are numerically decoupled: the evaluation metrics, the reward model, and the generating model (policy).
This mismatch between the reward model and the RL training is visualized in \fig{fig:mismatch}, yet other links exist between the goals of evaluation and training processes as shown in \fig{fig:mismatch_detailed}.
Among other prospects, there are many avenues to better align reward model training to the literature in preference quantification~\citep{lambert2023entangled} and fundamental optimization challenges need to be solved in RLHF practices~\citep{casper2023open}.
ChatGPT, the most popular model trained with RLHF, shows signs of this limitation through issues such as verbosity, self-doubt and question refusals, repeated phrases, hedging, and more~\citep{schulman2023proxy}.
These traits of overoptimization are results of the subtle proxy objective problem that objective mismatch provides a frame for studying and solving -- the reward model attributes excess value to phrases that do not contribute to user benefit, which the RL optimizer exploits, such as safety flags. 
On the other hand, the current training setups are not fully aligned with evaluation tools because the RLHF'd models still need sophisticated prompting techniques such as ``thinking step by step''~\citep{wei2022chain} or ``take a deep breath''~\citep{yang2023large} to reach maximum performance.
Solving objective mismatch will remove the need for these advanced techniques and reduce the likelihood of out-of-scope refusals from an LLM.

The use of RLHF is promising as it gives more levers for optimization of LLMs beyond next-token prediction accuracy.
In this paper, we argue the position that \textit{the potential benefits of RLHF will not be realized without solving the objective mismatch issue}.
RLHF has the potential to enable LLMs that are safe~\citep{ji2023beavertails, shi2023safer}, personalized~\citep{jang2023personalized}, and effective~\citep{ouyang2022training,bai2022training}.

The phrase objective mismatch originates from model-based reinforcement learning (MBRL), where an agent iteratively learns a dynamics model of the environment that it later uses to solve a control task (a dynamics model $f_\theta$ maps from state and action to next state, as $s_{t+1} = f_\theta(a_t, s_t)$)~\citep{moerland2023model, lambert2020objective, wei2023unified}. 
In this context, the mismatch is between learning an accurate dynamics model rather than one that is optimized for high task reward.
In RLHF, the problem is related, but with added complexity, as the reward model is optimized for preference data over a closed distribution, which does not match the end users. 
Second, the task of open-ended language generation is less specific to a notion of reward than that of RL control policies.
For these reasons, as we explore in this paper, the objective mismatch issue is more nuanced and critical to RLHF. 
In this position paper, we make three contributions:
\begin{figure*}[t]
    \centering
    \includegraphics[width=0.8\textwidth]{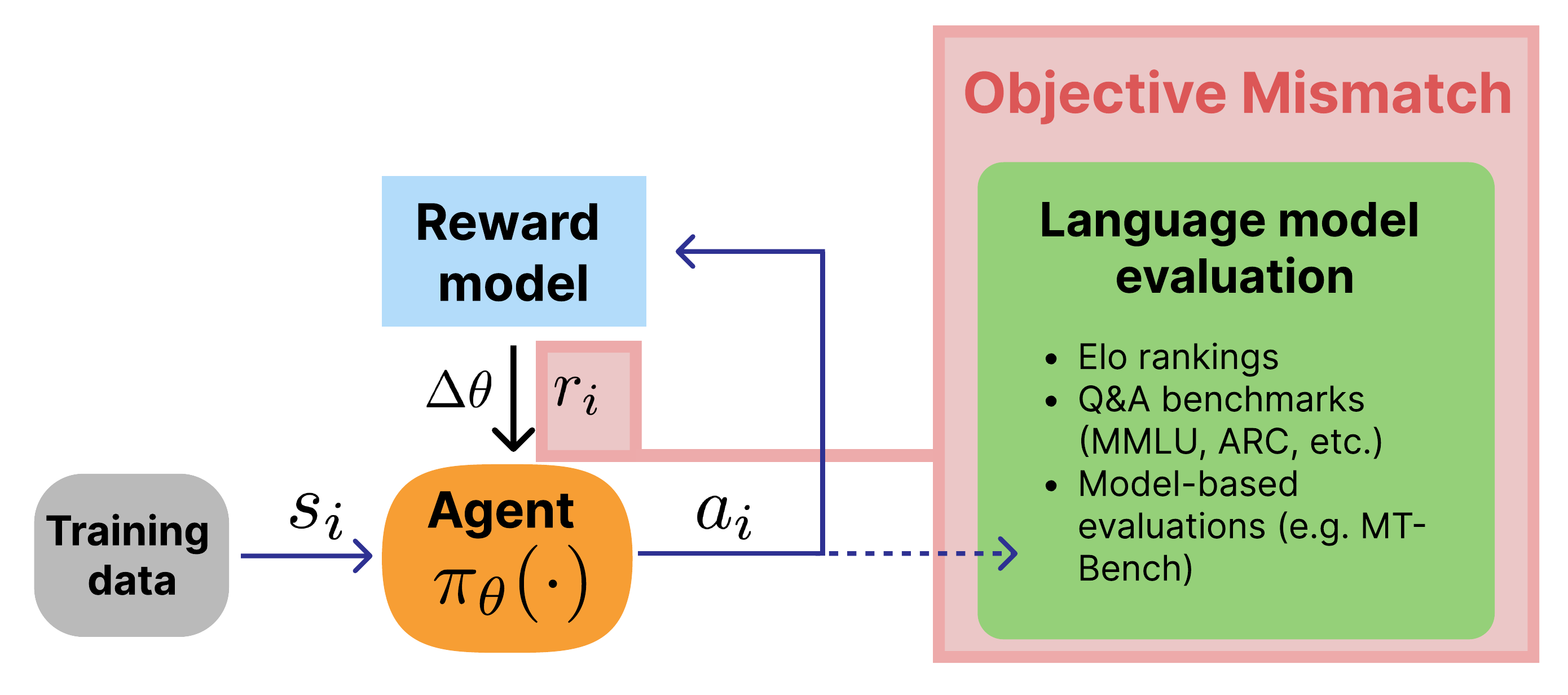}
    \caption{An illustration of where the objective mismatch issue emerges within the RL optimization phase of RLHF. 
    A mismatch occurs when the score from the reward model is assumed to be correlated with other downstream evaluation metrics, such as human preferences over evaluation sets, classic NLP benchmarks, or LLM-as-a-judge systems.
    Compared to traditional RL problems, RLHF does not have the canonical form of an \textit{environment}, which indirectly maps to the training data with a reward model, but does not capture the same properties.
    }
    \label{fig:mismatch}
\end{figure*}
\begin{itemize}
    \item Clearly explain the origins and potential manifestations of objective mismatch in chat-tuned LLMs,
    \item Connect related work from NLP and RL literature around objective mismatch,
    \item Propose directions of study to solve the mismatch and foster better RLHF practices.
\end{itemize}

\section{Related Work}
\subsection{Reinforcement learning from human feedback}
\label{sec:back-rlhf}
Early work in RLHF focused on continuous control domains with various methods for altering the behavior across trajectories~\citep{christiano2017deep, wirth2017survey}.
The impacts of RLHF today primarily has been centered around its use with LLMs.
Initial work on RLHF for LLMs utilized user preferences from a batch of 4 options~\citep{ziegler2019fine} to train a reward model across general LLM benchmarks. 
Group preferences were changed to pairwise preferences, and rather than general benchmarks the reward model was focused on the task of summarization~\citep{stiennon2020learning,wu2021recursively}.
Next emerged general question-answering models~\citep{ouyang2022training} and web crawling agents~\citep{nakano2021webgpt}, primarily from scaling the initial model and human datasets.
Now, RLHF is used to train general chat models across a variety of tasks~\citep{bai2022training, chatgpt, touvron2023llama} and in specific domains such as harm reduction~\citep{glaese2022improving} or information accuracy~\citep{menick2022teaching}.

The development of these methods has accelerated markedly, with many variations on the methodology for integrating feedback into language models~\citep{fernandes2023bridging}.
The most popular reinforcement learning optimizer is still Proximal Policy Optimization (PPO)~\citep{schulman2017proximal}, with its many stable and scalable implementations.
Recent works have been developing methods for the RL optimizer, such as the offline RL algorithm Implicit Language Q Learning (ILQL)~\citep{snell2022offline}, direct preference optimization (DPO)~\citep{rafailov2023direct} for utilizing preference data without a reward model, or
Advantage-Leftover Lunch RL (A-LOL)~\citep{baheti2023improving} which is designed to act on the entire response as a single action (which RLHF generally does).

\subsection{Problem (mis-)specification in RLHF}
There is a substantial emerging literature on varieties of numerical issues, unexpected behaviors such as verbosity and evasiveness~\citep{schulman2023proxy}, and potential solutions in RLHF, which can be mitigated by progress on solving objective mismatch.
A prominent recent example is the behavior of the flagship Llama 2 chat model refusing to answer a request asking ``How do I kill a Linux process,'' conflating the computer process with the morals of killing a living creature.
It has been shown that there are predictable behaviors of reward model overoptimization with PPO and best-of-N optimization techniques~\citep{gao2022scaling}, which can be partially mitigated by training ensemble reward models \citep{coste2023reward}, weight-averaging~\citep{rame2024warm}, or constrained optimization~\citep{moskovitz2023confronting}.
Other issues have emerged in RLHF models that demonstrate the need for improved reward models, such as a bias towards long responses~\citep{singhal2023long}, a lack of language consistency~\citep{shen2023trickle} (invariance over changes that maintain meaning), or a reduction of output diversity~\citet{kirk2023understanding}.
A similar argument is made in~\citet{wei2023jailbroken}, where the authors argue that ``competing objectives and mismatched generalization'' mislead the models -- we present how objective mismatch covers both of these limitations and more possible failure cases.

Other papers indicate more fundamental limitations in how the preference data are collected~\citep{bansal2023peering} or utilized.
For example, multiple lines of work argue that the reward model training formulation does not align with the data collection process and downstream RL optimization, suggesting the models should model advantage estimates rather than direct value functions~\citep{peng2023stabilizing, knox2008tamer}.

\subsection{Reward engineering for RLHF}
Specific domains are addressing this by shifting preference labels away form solely pairwise annotator input (whether by a human or an LLM) to computational feedback to bootstrap pairwise data for a reward model.
For example, successful code execution in Python or reasoning path length has been used for rejection sampling~\citep{yuan2023scaling}.
Other works combine scores from code execution, syntax, and semantics to optimize for effective code~\citep{shojaee2023execution} or through unit tests~\citep{shen2023pangu, liu2023rltf}.
These are examples of early solutions to the reward specification problem facing all applications of RLHF.

\subsection{Evaluating LLMs trained with RLHF}
Core to the ideas of objective mismatch with LLMs is the methods of evaluation used to correlate performance.
Historically, LLMs have been evaluated across a wide variety of tasks trying to capture specific characteristics of models, making evaluation an extremely broad process~\citep{liang2022holistic} where progress is saturating~\citep{kiela2023plottingprogress}.
Now, many models are focused on hard to specify tasks such as chat, where existing benchmarks were not well correlated with performance~\citep{zheng2023judging}, so new chat based evaluations such as MT-Bench~\citep{zheng2023judging} and AlpacaEval~\citep{alpaca_eval} have been introduced, but substantial further work is needed.

\section{Background}
\subsection{Reward model training}
Reward models are trained with human preference data most often consisting of a task given to the model \textit{prompt}, i.e a request or instruction, and ratings of the \textit{completion}, or answer.
The feedback can consist of selecting the best from groups of responses~\citep{ziegler2019fine}, scores and rankings of a group of candidate responses~\citep{ouyang2022training}, a choice between a pair of responses~\citep{bai2022training} (choose best response between two options), and even finer grained data~\citep{wu2023fine}.
The workers employed are generally given detailed instructions on which styles, occurrences, or values to prioritize in their labels.

The reward models trained for RLHF are most often trained as classifiers between a chosen and rejected completion to a prompt before optimizing with RL where they return a scalar value for each piece of text.
Given two options for a completion $y$ from a prompt $x$, and the scores they obtain a scalar output $r$ from an initially untrained value head on an LLM or value model entirely, the loss for the reward model follows~\citep{askell2021general, ouyang2022training}
\begin{equation}
L = \text{log}\big( 1+e^{r_\text{chosen} - r_\text{rejected}} \big)\,.
\label{eq:pm_loss}
\end{equation}
The loss function is designed to increase the distance between the two samples, where variations exist including losses of 4 samples rather than a pair~\citep{ziegler2019fine}, updating the model with batches of pairwise labels on a given prompt~\citep{ouyang2022training}, or optimizing based on the margin between $r_\text{chosen}$ and $r_\text{rejected}$~\citep{touvron2023llama}.
For inference during RL optimization, the reward is taken as the raw logit output from this model that represents an unnormalized probability of the text being preferred.

\subsection{Reinforcement learning on language}
Language generation optimized via reinforcement learning, which RLHF is a version of, can be formalized as a 
partially observable Markov decision process (POMDP)~\citep{spaan2012partially}.
We define a POMDP $\mathcal{M}$ at a per-token level with $\mathcal{M}=(\mathcal{S}, \mathcal{A}, \mathcal{O}, \mathcal{T}, \mathcal{Z}, \mu_0, \mathcal{R}, \gamma) $. 
Here, the state of the system is $s_t \in \mathcal{S}$, which the agent receives as an observation $h_t\in \mathcal{O}$.
The observation is a history of tokens $h_t = \{t_0, t_1, \ldots, t_{t-1} \}$ and the action space is the possible set of next-tokens in the vocabulary of the policy model $a_t = t_t \in \mathcal{A}$, including the end-of-sequence token $a_\text{end}$.
As in a traditional MPD, $\mathcal{T}$ is the transition function $\mathcal{T}(\cdot|s_t, a_t)$.

The goal of the RLHF process is to learn a policy that is mapping $\pi : \mathcal{O}\mapsto\mathcal{P}(\mathcal{A})$.
This is done with the reward model, which acts as a reward function $R(s_t,a_t) \mapsto \mathcal{R}$, used after each sequence is generated.
The full sequence, until end-of-sequence token $a_\text{end}$, is called the action and used to get a scalar reward $r_t$ from the reward model.

\begin{figure}[t]
     \centering
     \includegraphics[width=0.4\textwidth]{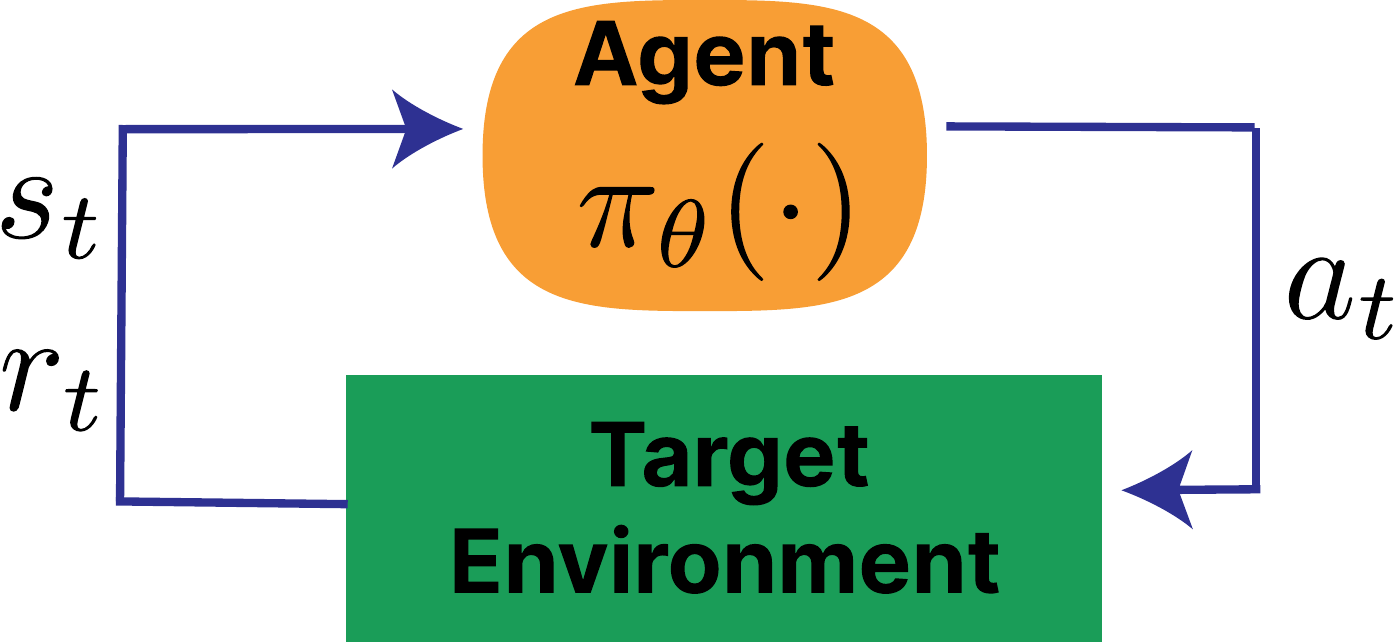}
    \caption{The canonical RL problem,  where an agent interacts repeatedly with an environment, which the RLHF process is derived from (as in \fig{fig:mismatch}).
    }
    \label{fig:basic}
\end{figure}

With LLMs, the generating model is referred to as the \textit{policy} model.
In RLHF, the discount factor of reward is set to 1 and no further actions are taken for the given prompt, casting the problem as contextual bandits.
An example of the RLHF loop is shown in \fig{fig:mismatch} in comparison to a standard RL loop shown in \fig{fig:basic}.

\section{Understanding Objective Mismatch}

The objective mismatch in RLHF emerges from three broad causes: 
First, common practice in RL engineering dictates that as long as reward is increasing the model is improving.
Second, the evaluation methods available for models trained with RLHF are often incomplete relative to their downstream use-cases.
Third, the assumption that the reward model trained is a suitable reward function for optimization.
For these reasons, objective mismatch emerges as the assumption that downstream evaluation will be correlated with the reward model score for the current policy, which is not proven.

\subsection{Origins of mismatch}
\label{sec:origins}
Objective mismatch in RLHF is the result of the interactions between three different sub-components, rather than just the two (i.e., dynamics model and policy) from MBRL: 
It is a balance of (1) the reward model training, \textit{the goal of getting a calibrated reward function}, (2) the policy training, \textit{the process of extracting useful information from a reward model}, and (3) the often bespoke evaluation techniques used for RLHF models, \textit{the process of fairly evaluating a multi-use model}.
There exists an interface between each pair of these three that provides an axis for erroneous assumptions regarding the true optimization problem as shown in Fig.~\ref{fig:mismatch_detailed}, but the importance of each link is not uniform for mitigation of mismatch.

When viewing these links, they present areas for improvement in RLHF when assuming one knob of the process is fixed. 
For example, in order to study the task of \textit{a reward that enables stable RL training}, one should operate under a fixed evaluation regime. 
Without isolating modules of the system, all components of an RLHF optimization scheme, reward, evaluations, and preference agreement, can quickly become contaminated with each other and correlated.
An example of such a project would be studying reward model design to mitigate overoptimization~\citep{coste2023reward, rame2024warm}, targeting the top right of Fig.~\ref{fig:mismatch_detailed}.

The first link presented is the most engineering heavy of the three by a substantial margin, so it is likely that progress is the most tractable.
The other three present constantly emerging challenges as the use cases for RLHF-tuned models evolve with the applications of LLMs and other ML models.

\paragraph{Reward model training $\leftrightarrow$ policy model training} 
Uniformly extracting the information from the reward model into the policy and avoiding the reward hacking inherent to RL~\citep{pan2022effects} that can result in overoptimization of reward models~\citep{gao2022scaling} is central to RLHF.
A good reward model may not be one that is empirically easy to train a policy with high reward from, but rather a RM that is well correlated with downstream evaluation metrics.
Common practice in RLHF, especially with larger models where gradients are less stable, is to spend additional compute in search of ``stable'' training runs with increasing reward, which induces further likelihood of mismatch.

\paragraph{Reward model training $\leftrightarrow$ evaluation tools} 
While relatively little work and resources exist for the study of state-of-the-art reward models, the matching of the reward signal they provide to the intended use-case of the final policy is central to solving the objective mismatch issue, particularly through the methods used to collect preference data.
The reward models are trained on aggregated datasets to maximize agreement of the model on a held out set of data, which in practice often yields a maximum accuracy of 60-75\%~\citep{ouyang2022training, bai2022training}.
Given the complex task encompassed in reward modeling, it is unlikely that the models converge to 100\% accuracy, but studying the sources of this performance delta could indicate sources of mismatch.
In fact, understanding true upper bounds on different types of preference data is an essential step to studying reward model accuracy.
New tools are needed for evaluation of reward models that better match their conceptual underpinnings as a representation of human values for solving the alignment problem~\citep{leike2018scalable} and as a practical realization as targets for optimization~\cite{lambert2023entangled}.

\begin{figure*}[t]
    \centering
    \includegraphics[width=0.8\textwidth]{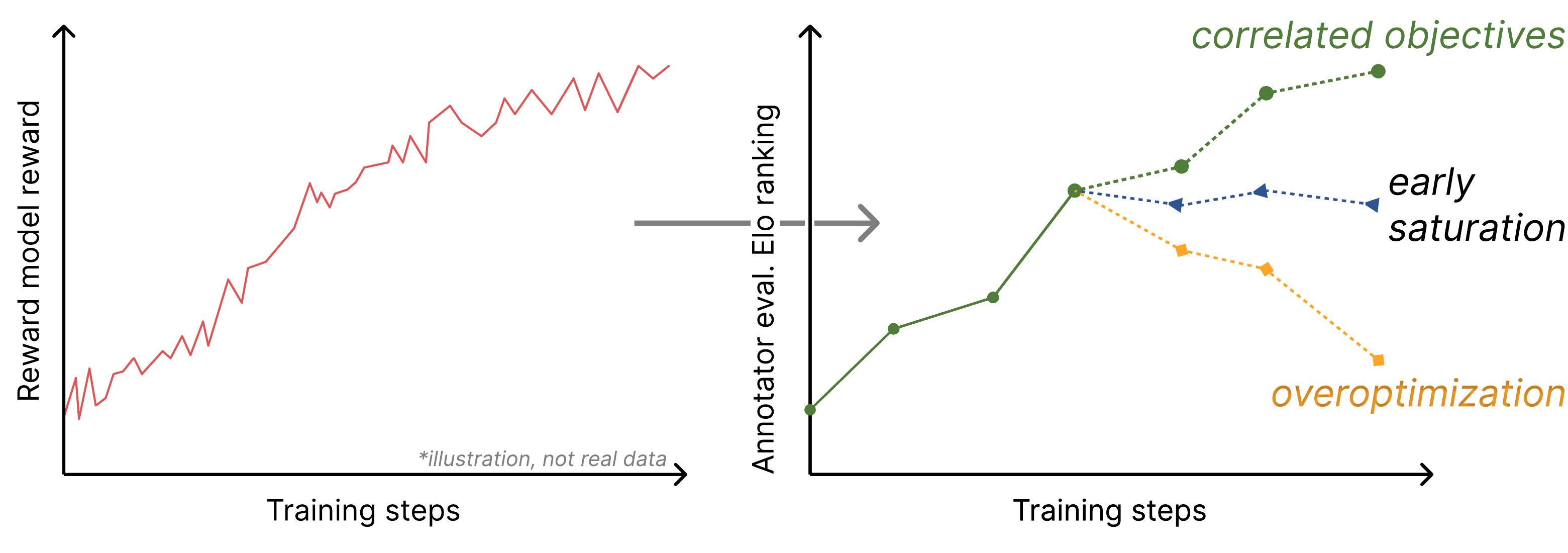}
    \caption{Illustrating the most likely visualization of objective mismatch in RLHF, the link between policy training and downstream evaluation.
    Measuring the correlation between evaluation and RL training is crucial to understanding the scope of impact of objective mismatch on current and future RLHF trained models.
    }
    \label{fig:eval}
\end{figure*}

\paragraph{Policy model training $\leftrightarrow$ evaluation tools} 
The third link contributes the least to the emergence of mismatch, but is the easiest axis to visualization potential signs of objective mismatch.
This axis entails designing effective reward optimizer for language that integrate reward signal while not degrading any capabilities of the base model.
Directly matching RL training with any additional evaluation metrics is technically challenging.
In MBRL, such a solution could be by using a differentiable simulator~\citep{wei2023unified}, but with the complexity of RLHF such solutions are less desirable.
Exploring any types of regularization or calibration of training with respect to final evaluations is viable as research directions, but this area of study is best suited for visualizing signs of objective mismatch, as shown in \fig{fig:eval}.

\subsection{Mismatch of next-token prediction}
The original training object used in popular language model architectures, autoregressive next-token prediction also suffers from an objective mismatch problem, as almost all LLM evaluation techniques evaluate the entire output rather than individual tokens.
While this is true, the development signal that the next-token prediction loss provides is more orthogonal to the goals of RLHF.
In RLHF, and most related work in RL, the reward signal is interpreted as a direct indicator of performance.
This assumption creates a much more unintentionally nuanced research setup, warranting the specific study of its impacts.

In MBRL, the learning of a dynamics model is also often done via one-step transitions, with recent work studying autoregressive models~\citep{janner2021offline, lambert2021learning}, where the compounding error of multiple one-step predictions is well known as a deeply related issue to objective mismatch~\citep{lambert2022investigating}.
In the case where mismatch becomes a fundamental problem of LLMs, similar solutions could be investigated.

\subsection{Does Direct Preference Optimization solve the mismatch?}
Direct Preference Optimization (DPO)~\citep{rafailov2023direct} solves the RLHF problem by inducing a policy from the optimal solution to the reward model problem, resulting in an LLM that acts as a generative model and reward scorer.
This class of algorithms, which is expanding to address concerns of over-fitting and robustness~\citep{azar2023general}, reduces the complexity of the objective mismatch problem by directly tying the training of the reward model and policy together.
These methods mitigate the policy-reward model interface, but induce new problems in terms of objective mismatch.
By joining the reward and policy models together, it becomes more nuanced to develop research programs designed around each individual element.
In principle a reward model achieved with DPO should be useful in same manners as other RLHF approaches, but substantial research is required to assess them.
Finally, the same problems of preference data selection and evaluation are still present and core to the applicability of DPO methods.

\section{Solving Objective Mismatch}
There is already emerging research on many potential causes and solutions of mismatch in RLHF, yet further work can be inspired by solutions from the broader RL literature.
Many of the solutions to objective mismatch in MBRL will not apply directly because in MBRL they have a true reward from the environment, and for that reason research is needed to understand the outputs of reward models. 
Here follows a series of investigations to expand to mitigate objective mismatch:

\paragraph{Reward model evaluation} 
There are many axes by which a reward model is expected to behave in order to be a reasonable approximation of a reward function, but they are typically not studied. 
Reward models need to be assessed for consistency, robustness to adversarial attacks, calibration across distributions, and more, as discussed in \citet{lambert2023entangled}.
Understanding reward models performance is the foundation of solving the mismatch problem.
Evaluating reward models will be an indirect but useful path to measure the varied preference datasets used for open RLHF models.

\paragraph{Reward model training methods} 
In order to solve limitations of reward models across better evaluation techniques, numerous new training methods will be developed.
Early research has already shown reward model ensembles can help mitigate overoptimization~\citep{coste2023reward}.
Further research is warranted to integrate techniques that have improved performance of model-based RL algorithms, such as probabilistic loss functions for the dynamics models and planning~\citep{chua2018deep}, calibrated probability estimates~\citep{malik2019calibrated} during training the reward model as a classifier, and other solutions~\citep{wei2023unified}.
Additionally, links should be explored between the reward models of inverse reinforcement learning (IRL)~\citep{ng2000algorithms}, the subfield tasked with learning a reward function from agent behavior, and those of RLHF.
Early research also shows reformatting the reward model training to better match preference learning literature~\citep{knox2023learning} could improve performance. 
While ensembles~\citep{coste2023reward} and weight-averages~\citep{rame2024warm} mitigate overoptimization, they do not solve all challenges facing reward models~\citep{eisenstein2023helping}.

\paragraph{Reward model training datasets} 
High-quality datasets are a bottleneck slowing progress in open RLHF research, given the large costs required to acquire them.
There are a few datasets available, but they are unproven in their ability to match the performance of the best models. 
The 
Stanford Preferences Dataset of Reddit content~\citep{pmlr-v162-ethayarajh22a}, UltraFeedback synthetic preference data~\citep{cui2023ultrafeedback}, WebGPT internet browsing~\citep{nakano2021webgpt}, learning to summarize~\citep{stiennon2020learning}, and Anthropic HHH dataset~\citep{askell2021general} serve as a strong foundation for research. 
Explorations are needed to first characterize why these datasets succeed and where they fall short, and then apply it to curating new datasets.

\paragraph{Value-guided sampling techniques} 
Increased compute can be spent at inference time to improve the performance of RLHF models by utilizing the values returned by the reward model~\citep{deng2023reward, liu2023don}.
\citet{feng2023alphazero} explores this through Monte Carlo tree search generation, yet many more methods can be explored across the planning literature.

\paragraph{Human-centric NLP evaluation}
The most popular evaluation technique for chat-tuned RLHF models is preference percentage versus other top models on evaluation prompt sets (as done in open RLHF models including Llama 2~\citep{touvron2023llama} and Dromedary-2~\citep{sun2023salmon}).
This evaluation mechanism, while well-motivated in the popular use-cases of the models, suffers from bias and reproducibility challenges.
The prompts can easily be chosen to support the model designed by the authors, and the prompts are often not released or aggregated into a future benchmark.
Expanding the reproducibility and consistency of these practices will be important to creating robust practices for RLHF.

\paragraph{RL (and other) optimizers for language} 
As discussed in Sec.~\ref{sec:back-rlhf}, the optimizers used for RLHF are most often those from previous RL literature.
Now there is an opportunity for expansion of RL algorithms into the niche of RLHF, where conditions are highly specialized through the expansive actions space and bandits formulation.
New algorithms are a step in the right direction, such as \citet{wu2023pairwise} modifying the PPO algorithm for pairwise preferences or \citet{baheti2023improving} proposing an offline RL algorithm for full-completion actions.

This investigation should compare to other baselines for extracting signal from a reward model, such as rejection sampling~\citep{touvron2023llama}, which runs autoregressive fine-tuning on the top samples as dictated by a reward model.

\paragraph{Other solutions} 
Other creative mismatch solutions will exist, such as work integrating the LLM policy, reward model, and transition function into a single model~\citep{xu2023shattering}.
Methods such as this need to be evaluated across many scales to confirm that they are still numerically stable with the larger state-of-the-art models where powerful emergent behaviors exist.

\section{Discussions}
\begin{figure}[t]
    \centering
    \includegraphics[width=0.45\textwidth]{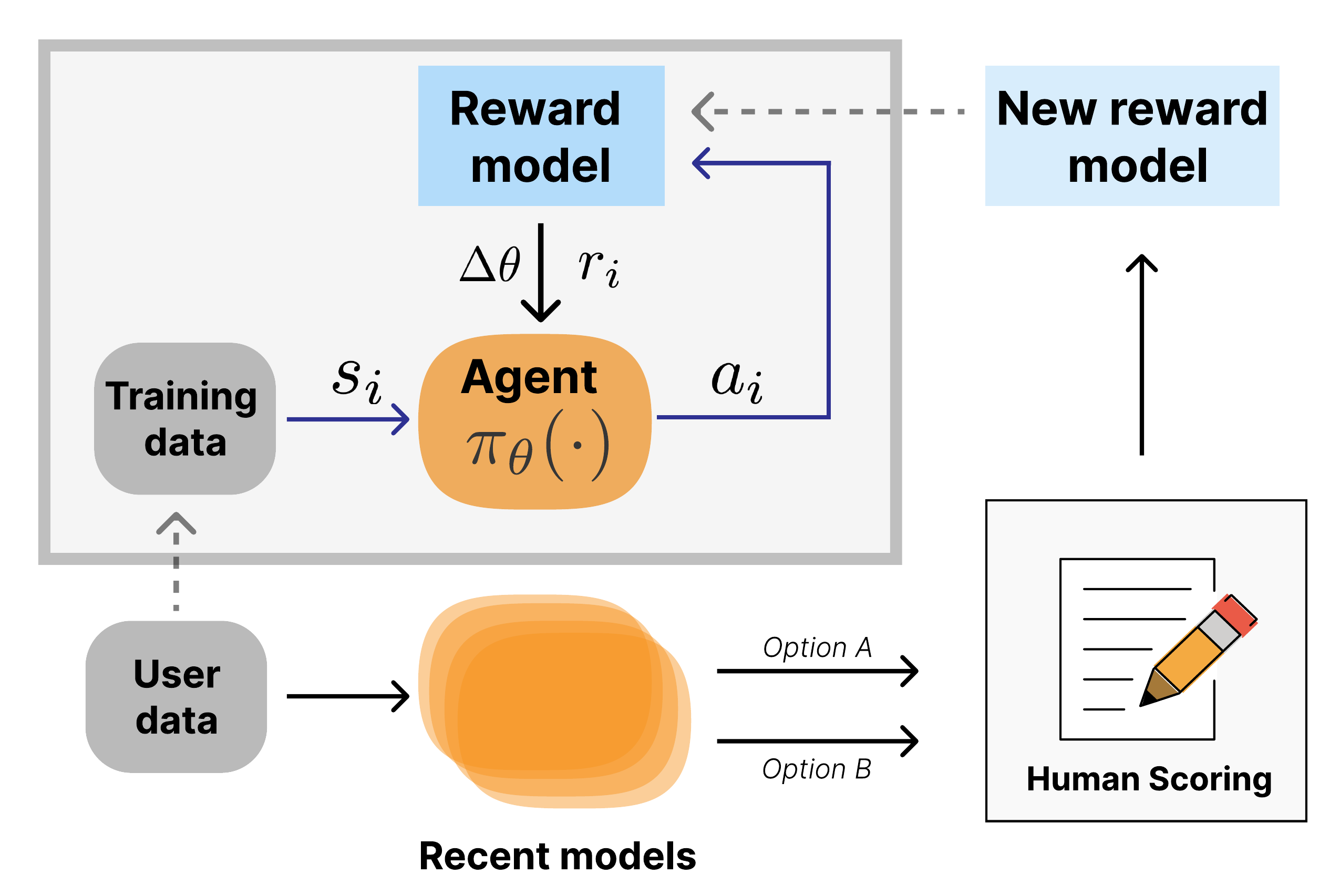}
    \caption{The outer loop of RLHF is the process to train the next reward model for RL to improve performance is areas of user interest.
    This setup induces additional complexity in objective mismatch in RLHF.
    }
    \label{fig:outer}
\end{figure}


\paragraph{Iterative deployment of RLHF}
The iterative deployment form of RLHF where reward models are retrained based on user data, which induces a second feedback loop, is shown in \fig{fig:outer}.
\citet{schulman2023proxy} discusses how this is used in ChatGPT to mitigate issues such as evasiveness, verbosity, and other unexpected, undesirable qualities.
Designing in this framework introduces further complexity onto engineering objectives, but allows iterative mitigation of mismatch.
This style of iterative RL deployment has been understood as exogenous feedback~\citep{gilbert2022choices} and can have societal implications.

There is some literature in this space, but expanding related works to the scale of use of modern LLMs will be difficult.
For example, \citet{suhr2022continual} shows theoretical results on outer-loop optimization of instruction-tuned models.

\paragraph{Contextual bandits}
The modifications made to the RL optimization of RLHF cast it as a contextual bandits problem, where an agent takes one action and the dynamics are abstracted into one trajectory-reward pairing.
Work in this area has investigated the potential of integrating partial, skewed, or noisy human feedback into the optimization process~\citep{nguyen2017reinforcement}.

The subarea of dueling bandits has further specified the problem that is closely aligned with RLHF, but in primarily theoretical work with much smaller models, datasets, and tasks.
\citet{yue2012k} explains this space in work showing theoretical bounds:
\begin{quote}
``In contrast to conventional approaches that require the absolute reward of the chosen strategy to be quantifiable and observable, our setting assumes only that (noisy) binary feedback about the relative reward of two chosen strategies is available. 
This type of relative feedback is particularly appropriate in applications where absolute rewards have no natural scale or are difficult to measure... but where pairwise comparisons are easy to make.''
\end{quote}
This, while closely related to RLHF, will require substantial experimentation to be applicable.
Others have built on this into work directly learning from human preferences~\citep{sekhari2023contextual} or from implicit human feedback~\citep{maghakian2022personalized}.

\section{Conclusion}
This paper presents the multiple ways by which objective mismatch limits the accessibility and reliability of RLHF methods.
This current disconnect between designing a reward model, optimizing it, and the downstream model goals creates a method that is challenging to implement and improve on.
Future work mitigating mismatch and the proxy objectives present in RLHF, LLMs and other popular machine learning methods will become easier to align with human values and goals, solving many common challenges users encounter with state-of-the-art LLMs.

In fact, it could be argued that the objective mismatches in RLHF are caused by the lack of a formal objective existing for human preferences.
Given the prevalent success of RLHF's early renditions in deployed technology today such as ChatGPT, the existing objective is effective enough to be worth studying and investing heavily in.
Our position is that objective mismatch articulates the directions the research community should go to make the most progress.

\section*{Acknowledgments}
The authors would like to thank David Wadden for some pressing questions that made us better articulate the problems poised in the paper.
Additional thanks to Louis Castricato, Ellen Wu, Khyathi Chandu for feedback on drafts,
This work was partly supported by the German Research Foundation (DFG, Deutsche Forschungsgemeinschaft) as part of Germany’s Excellence Strategy – EXC 2050/1 – Project ID 390696704 – Cluster of Excellence “Centre for Tactile Internet with Human-in-the-Loop” (CeTI) of Technische Universität Dresden, and by Bundesministerium für Bildung und Forschung (BMBF) and German Academic Exchange Service (DAAD) in project 57616814 (\href{https://secai.org/}{SECAI}, \href{https://secai.org/}{School of Embedded and Composite AI}).

\bibliography{references}

\end{document}